\documentclass{article}

\usepackage{PRIMEarxiv}

\usepackage[utf8]{inputenc} 
\usepackage[T1]{fontenc}    
\usepackage{hyperref}       
\usepackage{url}            
\usepackage{booktabs}       
\usepackage{amsfonts}       
\usepackage{nicefrac}       
\usepackage{microtype}      
\usepackage{lipsum}
\usepackage{fancyhdr}       
\usepackage{graphicx}       
\graphicspath{{media/}}     

\title{Virtual Scientific Companion for Synchrotron Beamlines: A Prototype 
}

\author{
  Daniel Potemkin \\
  Department of Physics \\
  Stony Brook University \\
  NY 11794, USA \\
   \And
  Carlos Soto \\
  Computational Science Initiative \\
  Brookhaven National Laboratory \\
  Upton, NY 11973, USA\\
  \And
  Ruipeng Li \\
  National Synchrotron Light Source II\\
  Brookhaven National Laboratory \\
  Upton, NY 11973, USA\\
  \And
  Kevin Yager \\
  Center for Functional Nanomaterials\\
  Brookhaven National Laboratory \\
  Upton, NY 11973, USA\\
  \And
  Esther Tsai\\
  Center for Functional Nanomaterials\\
  Brookhaven National Laboratory \\
  Upton, NY 11973, USA\\
  \texttt{etsai@bnl.gov} \\
}

\begin{document}
\maketitle

\begin{abstract}
The extraordinarily high X-ray flux and specialized instrumentation at synchrotron beamlines have enabled versatile in-situ and high throughput studies that are impossible elsewhere. Dexterous and efficient control of experiments are thus crucial for efficient beamline operation. Artificial intelligence and machine learning methods are constantly being developed to enhance facility performance, but the full potential of these developments can only be reached with efficient human-computer-interaction. 
Natural language is the most intuitive and efficient way for humans to communicate. However, the low credibility and reproducibility of existing large language models and tools demand extensive development to be made for robust and reliable performance for scientific purposes. 
In this work, we introduce the prototype of \underline{vi}rtual \underline{s}cientific compan\underline{ion} (VISION) and demonstrate that it is possible to control basic beamline operations through natural language with open-source language model and the limited computational resources at beamline.
The human-AI nature of VISION leverages existing automation systems and data framework at synchrotron beamlines. 
\end{abstract}


\section{Introduction}
\subsection{Synchrotron and Nano-science}
%

The extraordinarily high X-ray flux and coherence of modern synchrotrons have revolutionized the discovery and design of complex materials, enabling studies that cannot be pursued otherwise. X-ray imaging can, for example, reveal internal 3D nano-structure of integrated circuits~\cite{holler2017high, holler2019three}, battery cathode material~\cite{tsai2019correlated}, and human neurons~\cite{shahmoradian2017three, tran2020alterations}. X-ray scattering on the other hand can characterize bulk or single crystalline photovoltaics materials~\cite{li2021light}, polymer self-assembly~\cite{smilgies2022gisaxs, leniart2020large}, and organic thin films~\cite{xia2021solvent, levine2020crystal} by providing statistical information on nano to atomic structures, morphology, orientation, or crystallinity. In general, the design and understanding of functional materials often entail exploring a large library of materials with varying parameters (e.g., molecular weights, number and type of components, or treatment processes), the observation of in-situ structural or morphology changes under external stimuli, or the study of spatial/temporal/spectral changes of specific features. These studies are often only feasible at synchrotrons: compared to lab scale benchtop X-ray sources, synchrotrons provide a much higher photon flux, higher energy resolution and coherence, and often smaller beam spot sizes. These advantages directly enable versatile characterization methods, high throughput acquisition, and in-situ or operando studies. Synchrotron beamlines are therefore often oversubscribed and understaffed. Every year tens of thousands of users around the world are utilizing the unique resources and expertise at synchrotrons. 
%
With synchrotron upgrades around the world, data collection rates have increased even more: orders of magnitude more experiments can be performed within the same time frame, leaving beamlines understaffed and operation and analysis protocols inefficient. While scientific discussions should lead the user collaborations, staff are regularly preoccupied by technical and routine operations, including simultaneously controlling beamline hardware, acquisition, and analysis software for extended hours. 
On the other hand, the user community presents a broad range of expertise and background, e.g., some users have extensive experience in X-ray characterization and strong programming skills, while some users focus on material synthesis and the design of in-situ apparatus. 
To attract and sustainably support users from a diverse background, staff often devotes extra effort to guide users on all aspects of the experiment. In this work, we introduce the prototype of virtual scientific companion (VISION)~\cite{ECA_Tsai_2023} based on natural language processing (NLP) and large language models (LLMs) to enable NL-based interaction for efficient and intuitive experimentation.

Recent developments are being made to apply advances in LLMs to chemistry and material design~\cite{boiko2023autonomous, jablonka2023leveraging, white2023assessment}. Moreover, Yager showed that existing LLM and software tools can be easily adapted to build a domain-specific chatbot~\cite{yager2023domain}. Prince et al employed GPT-3.5 and open-source model Vicuna to illustrate the applicability of LLMs on experiment guidance~\cite{prince2023opportunities}. Advances of LLMs for scientific applications have the potential to start a new era where NL-based communication will be the only interface that scientists need for complex experimentation while other software/applications will be the backends of the NLP methods. 

\subsection{Natural Language and Language Models}
Many new artificial intelligence and machine learning (AI/ML) methods are being developed to address varying aspects of beamlines, including beamline optics, operation, or analytics. However, few methods have lasting impact on beamline experimentation due to limited communication between human and AI systems for dynamic interactions, as well as bottlenecks in knowledge transfer, deployment, and software maintenance and upgrades. The lack of a common language between different domain scientists—and between human and computer—remains a rift to be filled. 
Over the past year, OpenAI’s chatGPT chatbot has demonstrated the impressive ability to engage in informative and humorous conversations, ask follow-up questions, self-correct based on subsequent dialogues, write creative and summary essays, and even write and debug code. 
Smart voice assistants have become popular thanks to mature automatic speech  recognition models, e.g., speech models wav2vec~\cite{scheinker2020adaptive, baevski2020wav2vec} and Whisper~\cite{radford2022robust} are able to recognize multiple languages and dialects. The use of NL-based AI can be the solution to communication barriers between researchers from various backgrounds and between human and computer. 

While AI assistants’ creativity and assumed ability to summarize and consolidate massive information are impressive, the credibility and authenticity of the created content are in general rather difficult to be verified. Simple calculations are often not processed based on mathematical operations but rather generated and justified by seemingly correct text descriptions.
Furthermore, the non-deterministic nature (same input can trigger different responses) of chatbots can be perceived as creative and interesting when it comes to casual conversations or when creativity is sought, but for scientific discussions and experimentation that require accurate information extraction and exchange, reproducibility and credibility are essential. 
The low reliability of existing chatbots or voice recognition calls for further research and development to be made for virtual assistants for scientific purposes. Our vision is to build a framework to enable NL-controlled scientific expedition with joint human-AI force for accelerated scientific discovery. The current blooming development in NLP and AI chatbots provide abundant resources and tools for developing domain-specific applications for X-ray and nanoscience. 
Research on novel domain-specific LLMs could lead to scientific contextual understanding as well as the development of NLP systems for robust NL-based interaction. 
Nonetheless, large language models are not all open-source or free of charge; domain-specific development still needs to be made to address issues with existing technology. For research and national security reasons, NLP platforms for certain scientific exploration should to be developed and deployed on local network. 

Language models are generated by pre-training on a huge corpus of unlabeled text to leverage the linguistic information in NL text; that is, to understand the syntax and semantics of the language. Recent LLMs include, for example, the Generative Pre-trained Transformer (GPT)~\cite{brown2020language, radford2018improving, radford2019language, openai2023gpt4} series by OpenAI, Bidirectional Encoder Representations from Transformers (BERT)~\cite{devlin2018bert}, Language Model for Dialogue Applications (LaMDA)~\cite{thoppilan2022lamda}, and Pathways Language Model (PaLM)~\cite{chowdhery2022palm} by Google, Chinchilla~\cite{hoffmann2022training} by DeepMind, Galactica (GAL)~\cite{taylor2022galactica} and Large Language Model Meta AI (LLaMA)~\cite{touvron2023llama} by Meta, Mistral~\cite{jiang2023mistral} and Mixtral by Mistral AI, and many more. These models vary in terms of architecture (left-to-right relies on only previous words for prediction, while bidirection learns the context), training strategies, training data (public/restricted/domain-specific text or code corpus), model size (millions or billions of parameters), tokens (different subwords as basic units), and specialized NLP tasks and corresponding performances. NLP tasks, e.g., question and answering, classification, and named entity recognition (NER), can be performed by some LLM directly or through fine-tuning. 
Despite the ability for LLM to perform various natural language tasks, these models are still susceptible to hallucinations where generated text is incorrect or nonsensical. Nonetheless, combined with rule-based methods, it is possible to utilize LLM's ability for information retrieval to operate scientific instruments.  

\section{Methods and Results}

\subsection{LLM for Named Entity Recognition}
\begin{figure}[b]
  \centering
  \includegraphics[width=0.85\linewidth]{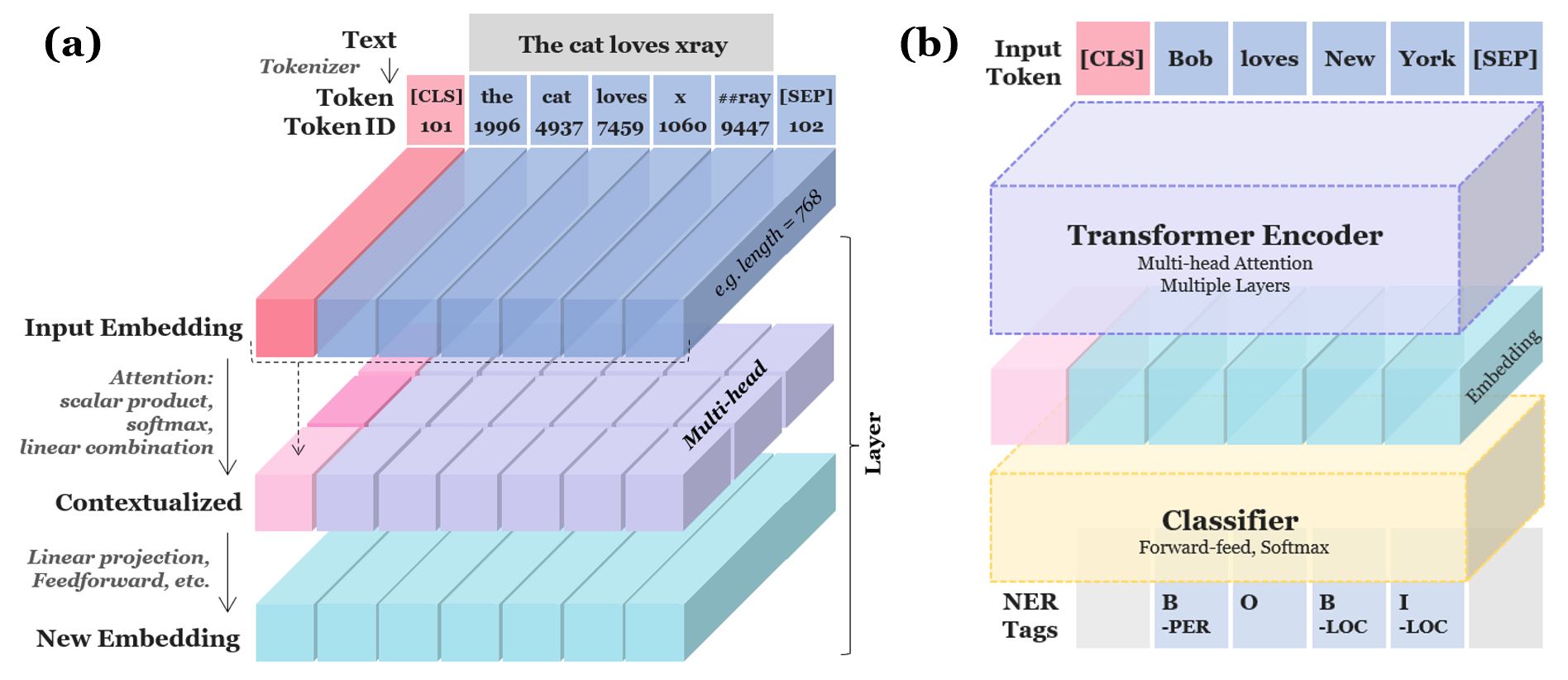}
  \caption{(a) In a Transformer, a sentence is split into tokens where each token corresponds to an embedding vector. Multi-head attention and multiple layers are applied to generate contextualized embeddings. 
  (b) Specific tasks can be achieved by fine-tuning BERT. In NER, each token is tagged with entity information.}
  \label{fig:LLM}
\end{figure}

In a Transformer, a sentence is broken into tokens through a tokenizer~\cite{wu2016google}, which employs a specific strategy to split a word into sub-words, i.e. tokens. The use of tokens allows new vocabularies unknown to the language model to be represented, that is, limited number of tokens can express unlimited words. As illustrated in Fig.~\ref{fig:LLM}(a), each token is represented by an embedding vector, the position of each token by a position embedding, and each sentence by a segment embedding. These embedding vectors are added together to be used as input to the Transformer, where self-attention~\cite{vaswani2017attention} is applied via cosine-similarity or correlation between each token to generate contextualized embeddings. In multi-head attention, embeddings go through different linear projections to give different contextualized vectors, which are then concatenated together, as illustrated by the middle purple blocks in Fig.~\ref{fig:LLM}(a). The same architecture can be used in pre-training from a large corpus or fine-tuning for specific downstream tasks. 
As illustrated in Fig.~\ref{fig:LLM}(b), the NER model processed the user input and tagged each word using the BIO tagging system, where ‘B’ was for the beginning of the keyword (or entity), ‘I’ for inside the keywords, and ‘O’ for outside, meaning the word/token is not of interest. Once these keywords were identified from the NL text, the beamline-interface utilized simple logic (if/case) to produce commands/scripts based on the keywords. 

In this work, we utilized BERT models available on the AI platform Hugging Face~\cite{huggingface_bert, wolf2020huggingfaces} to fine-tune the base model for NER. Models are trained to identify in a natural-language sentence the pre-defined keywords, such as person names, locations, or beamline-specific information such as temperatures, exposure time, or measurement geometry. 
The BERT model was used at the time of this implementation; many newer and more powerful models have been developed since. Here we showed that even with a basic BERT model, it is feasible to perform experimental control of scientific instrumentation. 
The BERT base model, available on the AI platform Hugging Face, was pre-trained on BooksCorpus (800M words) and Wikipedia (2500M words) with 12 multi-heads and 12 layers of transformers. Masked language model (MLM)~\cite{taylor1953cloze} and next sequence prediction (NSP) were the two objectives used in BERT pre-training of this large corpus of unlabeled text to enable deep bidirectional representation: in MLM, 15$\%$ of the words were removed or replaced with incorrect words; in NSP, sentences with the incorrect order were used to train the model. Given this base model, we fine-tuned it to retrieve beamline-specific information from natural-language input.

To generate large data sets for fine-tuning BERT, we listed dozens of sentence templates in different categories and randomly concatenate these sentences together to form a short paragraph. In total there are around 150 sentence templates, which is still a reasonable quantity to list manually or with assistance from LLM for paraphrasing. The sentence templates used in the training and test datasets are different: 80$\%$ of the templates in each category are used for training, the rest 20$\%$ for testing.
Table~\ref{table_data} listed the categories and selected sentence templates. For example, there are sentences related to changing the hardware settings, including sample stage temperature or position, or the measurement parameters or conditions. 
Keywords or entities in the template were filled randomly from a list of words or numbers, given in Table~\ref{table_key}. Each keyword is appended with “B-” or “I-” for the beginning or inside of a command. This distinction is used for logic processing for generating specific commands for instrument control. The model was trained to recognize 40 keywords. 

\begin{table}
 \caption{Data-generation by randomly concatenating sentence templates from different categories}
  \centering
  \begin{tabular}{p{0.15\linewidth} | p{0.85\linewidth}} 
    \toprule
    Category     & Sentence template (selected)   \\
    \midrule
    Hardware  & $\bullet$ Set the temperature to [B-TEMPERATURE] degrees at a rate of [I-NRAMP-MIN] degrees per minute \\
            & $\bullet$ Change humidity to [B-HUMIDITY] \\
             & $\bullet$ Move motor y by [B-YPOS-REL] \\    
    Parameter     & $\bullet$ Using incident angle [B-ANGLE] with exposure time of [I-ETIME] seconds   \\
                    & $\bullet$ Use incident angles [B-ANGLE] and [I-ANGLE] with exposure time of [I-ETIME] seconds \\
    Measure     & $\bullet$ Take a [B-SCAN] on the sample at [I-POINT-ABS]   \\
                & $\bullet$ Using [B-PROCESS], [I-SCAN] sample across [I-DIRECTION] \\
    Condition    & $\bullet$ Increase temperature until the temperature reaches [B-TEMPERATURE-CONDITIONAL] using rate [I-NRAMP-SEC] degrees per seconds and do the following every [B-TRATE-SEC] seconds \\
                & $\bullet$ Do this for [B-AMOUNT] times every [I-TRATE-SEC] seconds \\
    \bottomrule
  \end{tabular}
  \label{table_data}
\end{table}

\begin{table}
 \caption{Examples of keywords (entities)}
  \centering
  \begin{tabular}{p{0.15\linewidth} | p{0.85\linewidth}} 
    \toprule
        Keyword (Entity)     & List of words or numbers    \\
    \midrule
    SCAN & "exposure", "scan", "picture", "snapshot", "measure", "measurement", "look at", "see" \\
    PROCESS & "GISAXS", "TSAXS", "GIWAXS", "TWAXS" \\
    ETIME & Exposure time, e.g. np.arange(1, 200, step=1) \\ 
    XPOS-REL & Motor x relative position, e.g. np.arange(-100,100, step=0.1)\\
    TRATE-SEC  & Measurement period, e.g. np.arange(1, 200, step=1) \\
    TEMPERATURE & Temperature, e.g. np.arange(-200, 600, step=0.1) \\ 
    NRAMP-MIN  & Temperature ramp rate, e.g. np.arange(1, 200, step=1) \\
    
    \bottomrule
  \end{tabular}
  \label{table_key}
\end{table}

The randomness in the number of sentence templates and their orders as well as the randomly filled-in words for each keyword allows for easy generation of large amount of data. The generated short paragraphs might be grammatically awkward or not realistic for experiments, but allows us to generate large quantity of data very quickly: 10 seconds for generating 10k short paragraphs. In each short paragraph there were around 25 $\pm$ 8 keywords. 
Table~\ref{table_result} provides examples of short paragraphs and its labels or entities recognized by the BERT-NER.
Table~\ref{table_accuracy} provides the NER results and training time. For fine-tuning bert-base-uncased with 10k paragraphs and one epoch, it took 4 minutes on a single NVIDIA GeForce RTX 3060; the inference for 2000 entries (short paragraphs) took less than 20 seconds. 
The NER results for accurately recognizing the entity of a token is almost $99\%$, accuracy for identifying B- and I- entities that are important for formulating downstream commands is roughly $97\%$, and the paragraph accuracy is $70{-}75\%$ when accuracy is defined as all tokens in the paragraph are labelled correctly. For experiment control, paragraph accuracy is important to ensure robust NL-text interpretation and task execution. 
%
Varying the data size or BERT model does not show obvious change in accuracy, which could be attributed to the model's ability to quickly learn the NER task, the effects of training parameters, and/or the limitation in data generation.     

\begin{table}
 \caption{Short paragraph with its recognized entities}
  \centering
  \begin{tabular}{p{0.9\linewidth}} 
      Text [Predicted Entity]  \\
    \midrule
    $\bullet$ Scan [B-SCAN] across the x direction [I-DIRECTION] with GISAXS [B-PROCESS] on the sample using incident angle 0.19 [I-ANGLE] with exposure time of 10 [I-ETIME] seconds, do this for 10 [I-AMOUNT] times every 2 minutes [I-TRATE-MIN] \\  
     $\bullet$ Increase the temperature to 200 [B-TEMPERATURE] at 20 [I-NRAMP-MIN] degrees per minute, take a measurement [I-SCAN] of the sample at (2, 3) [I-POINT-AB] for 30 [I-ETIME] seconds at angles 0.1 [I-ANGLE], 0.29 [I-ANGLE], do this until the temperature is 400 [B-TEMPERATURE-CONDITIONAL] degrees at a rate of 20.2 [I-NRAMP-MIN] degrees per minute \\    
      $\bullet$ Move the motor x by 0.2mm [B-XPOS-REL] and measure for 1 [I-ETIME] second every 20 [I-TRATE] seconds until temperature reaches 300 [B-TEMPERATURE-CONDITIONAL] deg with a rate of 10 [I-NRAMP degree per minute  \\   
      $\bullet$ Measure [B-SCAN] this polymer [B-SAMPLE] sample Change the temperature to 300 [B-TEMPERATURE] degree and measure [I-SCAN] for 1 [I-ETIME] second every 60 [I-TRATE-SEC] seconds  \\    
    \bottomrule
  \end{tabular}
  \label{table_result}
\end{table}

\begin{table}
 \caption{Data size and NER accuracy}
  \centering 
  \begin{tabular}{p{0.08\linewidth} | p{0.08\linewidth} | p{0.17\linewidth}  | p{0.2\linewidth}   | p{0.2\linewidth}  | p{0.07\linewidth}} 
    \toprule
        Model & Data size  &  Paragraph & Token (all) & Token (B-, I-, ) & Training   \\ 
        ($\#$param) & (training 0.8) &  Accuracy & Accuracy & Accuracy & time (min.)\\ 
    \midrule
  base &      15000 & 0.743 (2230/3000) & 0.985 (162173/164606) & 0.970 (74705/77018) & 6.9\\ 
  (110M)     & 10000 & 0.755 (1511/2000) & 0.987 (106507/107884) & 0.975 (49393/50679) &  4.4 \\
        & 5000 & 0.722 (722/1000) &  0.985 (53529/54318) & 0.970 (24634/25386) & 2.2\\
    \midrule
large &    15000 &  0.693 (2080/3000)  & 0.983 (161765/164606) & 0.965 (74359/77018) & 22.5\\
(340M)        & 10000 &  0.735 (1469/2000) &  0.986 (106387/107884) & 0.973 (49288/50679) & 14.8\\
              & 5000 & 0.763 (763/1000)  & 0.988 (53663/54318) & 0.976 (24772/25386) & 7.3\\ 
    \bottomrule
  \end{tabular}
  \label{table_accuracy}
\end{table}


\subsection{NER for Beamline Operation}
Existing Python-based infrastructure at NSLS-II beamlines allows beamline components to be easily modified and precisely controlled. However, synchronizing multiple components, e.g., concurrent acquisition, analysis, and autonomous algorithms, pre-occupies one or more staff scientists during hectic beamtimes. 
Data acquisition at NSLS-II beamlines is performed under a Python environment through the Bluesky data collection framework~\cite{bluesky}, established by the Data Science and Systems Integration (DSSI) program at NSLS-II. 
Real-time batch analysis of X-ray scattering data are usually provided by customized beamline-specific Python program.
These Python-driven operations at the beamline provide the foundation for advanced AI/ML developments. 

\begin{figure}
  \centering
  \includegraphics[width=0.43\linewidth]{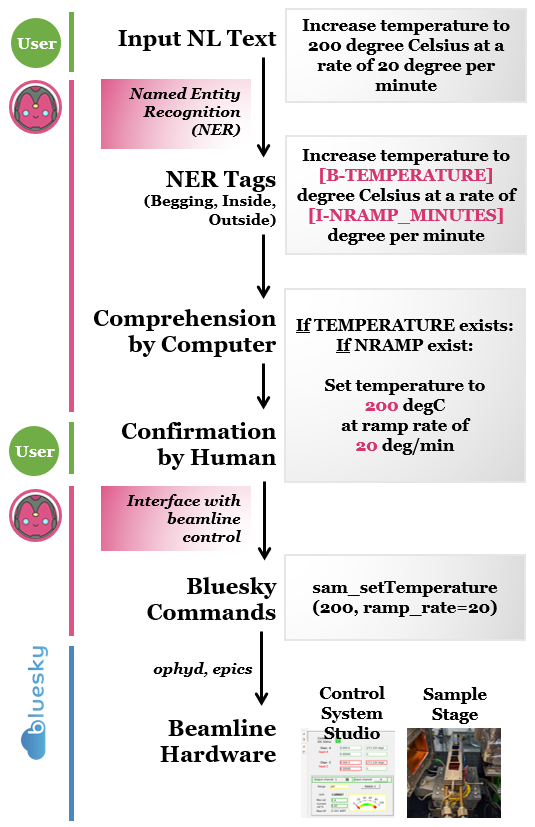}
  \caption{Pipeline for converting NL-text to commands for beamline experiment control. 
  }
  \label{fig:pipeline}
\end{figure}

\begin{figure}
  \centering
  \includegraphics[width=0.9\linewidth]{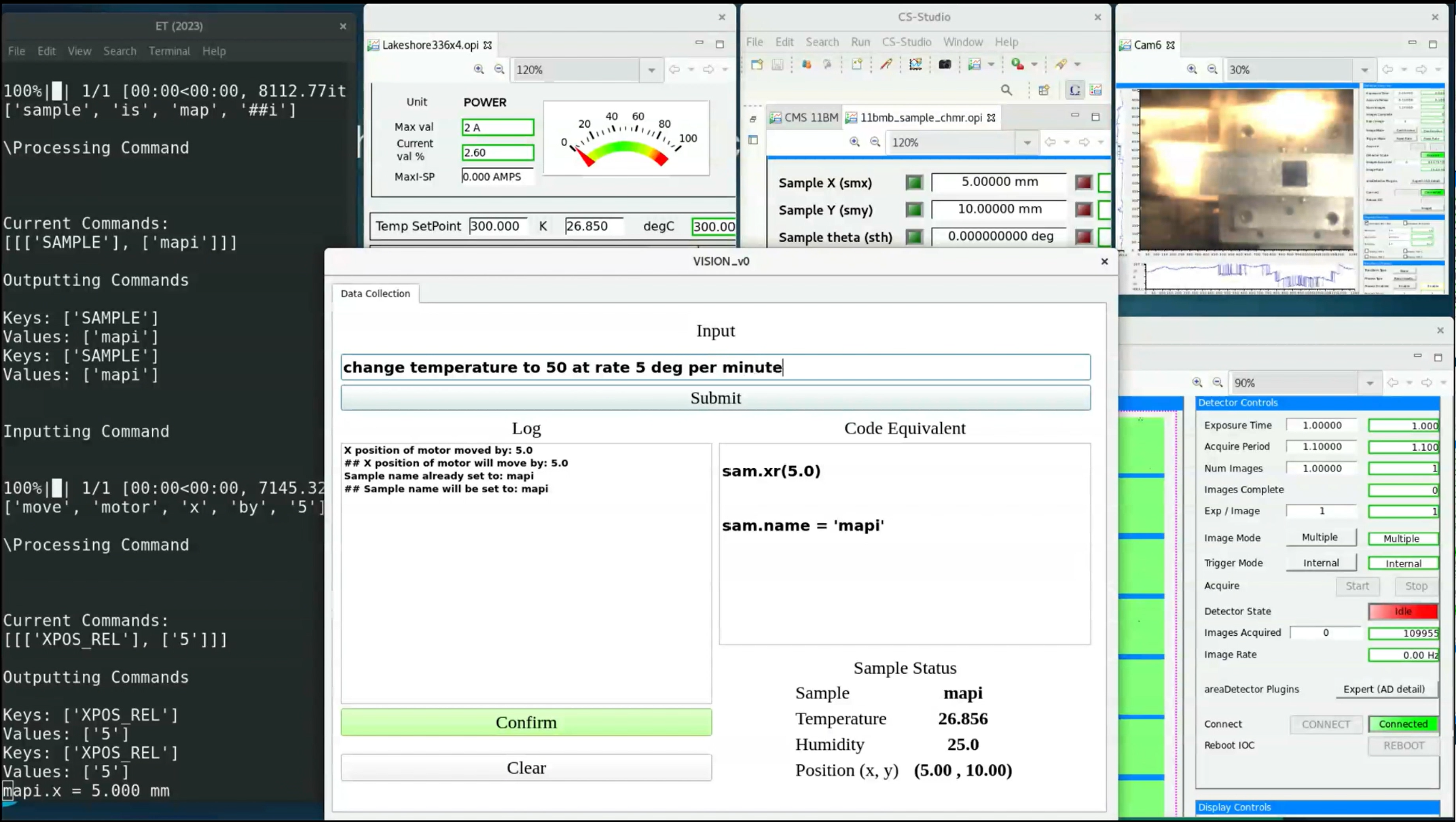}
  \caption{VISION prototype deployed at NSLS-II beamline 11-BM CMS, showing (left) the backend, (middle) the UI, and (top and right) the control interfaces for beamline hardware. See~\cite{VISION_demo} for video. }
  \label{fig:GUI}
\end{figure}

We demonstrated the use of BERT-NER for extracting information from NL descriptions for downstream beamline data acqusition at NSLS-II X-ray scattering beamline11-BM CMS. 
VISION prototype consists of three main parts: (i) a {backend} that employs the NER model and extracts keywords for specifying downstream operation and parameters, (ii) an {interface class} that used logic or rule-based methods to converts keywords into Bluesky commands or scripts for experiment control, and (iii) a simple {graphical user interface} (GUI) for users to enter requests in NL-text, confirm interpretation, see the code equivalent, and execute the task. Python commands can also be entered through the GUI input.
In the example given in Fig.\ref{fig:pipeline}, the entities were the target temperature of the sample and the heating ramp rate to be used to achieve the target temperature. The interpreted result was displayed, prompting the user to confirm action. The confirmation step was crucial to prevent unwanted changes on the sample condition or beamline hardware. Once the action was confirmed by the user, the code equivalent to the NL input was displayed to the GUI and executed on BlueSky to control beamline instrumentation. Figure~\ref{fig:GUI} shows the backend, the GUI, and the control interface for beamline hardware. A demo video can be found at~\cite{VISION_demo}, which shows the use of NL-text for sample naming, motor movement, temperature change, parameter change, and data collection.

\section{Conclusion}
Here we demonstrated that existing low-budget resources can be easily utilized for experiment control at synchrotron beamlines:

\begin{itemize}
\item Provide scheme for quick data generation 
\item Fine-tune open-source basic language model for NER for beamline operation 
\item Demonstrate LLM's compatibility with local beamline workstation with limited computation power (NVIDIA GeForce RTX 3060)
\item Show reasonable inference time (few seconds)
\item Present great potential of LLM for complex beamline operations, customized experimentation, and efficient human-AI interaction for specialized instrumentation in general.
\end{itemize}

In this work, the BERT model in the early LLM development was used to demonstrate the feasibility of using LLM for experiment control. With the rapid development in LLMs, prompt engineering can be utilized for information-retrieval without fine-tuning or create a more interactive environment with chat history. Advanced LLMs can be incorporated to provide better reasoning and acting~\cite{yao2023react} or utilize existing tools to enhance reliability~\cite{schick2023toolformer, gao2023pal}. To fine-tune LLMs for downstream task or for adding domain-knowledge, data generation can be performed by retrieving information from existing documentations (e.g. beamline proposals, publications, related website, etc) in addition to or in lieu of sentence templates. 
For practical applications, however, the large size of some LLMs may require quantization~\cite{frantar2023gptq, GGML, dettmers2023case}, parameter-efficient fine-tuning~\cite{liu2022few, hu2021lora}, or more powerful computational resources.

On-going development continues at synchrotron beamlines for building closed-loop workflow with in-situ automated material synthesis/processing, X-ray characterization, data analysis, and integrated human-AI decision making for revolutionary studies of nanomaterials.
Autonomous experimentations via the gpCAM software~\cite{noack2021gaussian, noack2021gaussian, doerk2023autonomous} have been demonstrated at NSLS-II beamlines and become part of the beamline capabilities for general users. 
We envision that in the near future virtual assistants can lead the NL-controlled scientific expedition with joint human-AI force for accelerated scientific discovery: 
they can provide a NL-based interactive framework with essential roles as an operator to acquire data, an analyst to process and visualize data,a transcriber to transcribe NL voice to text, and a tutor to provide plug-in functionalities, including advanced learning algorithms and physics and computation modelling. 
This work was specifically tailored for demonstrating LLM for experiment control at X-ray scattering beamlines, with impact however beyond synchrotron beamlines but also on nanocenters or any multi-tasking scientific or engineering workstation. 
The rapid development in NLP among private sectors provides many tools to be utilized and abundant experience to learn from, while also highlighting the need for dedicated research effort in the scientific community to integrate and adapt the NLP technology for scientific discovery.


\section*{Acknowledgments}
This project was supported in part by the U.S. Department of Energy, Office of Science, Office of Workforce Development for Teachers and Scientists (WDTS) under the Science Undergraduate Laboratory Internships Program (SULI). The work was supported by the DOE Early Career Award 2023.
This research also used beamline 11BM (CMS) of the National Synchrotron Light Source II and the Center for Functional Nanomaterials (CFN), both of which are U.S. Department of Energy (DOE) Office of Science User Facilities operated for the DOE Office of Science by Brookhaven National Laboratory under Contract No. DE-SC0012704. 

\bibliographystyle{unsrt}  
\bibliography{NLP, xray}

\end{document}